\documentclass[twoside]{article}
\usepackage[accepted]{aistats2019}

\usepackage[T1]{fontenc}    % Use 8-bit T1 fonts

\usepackage[round]{natbib}

\usepackage{microtype}      % Microtypography
\usepackage[utf8]{inputenc} % Allow utf-8 input
\usepackage{hyperref}       % Hyperlinks
\usepackage{url}            % Simple URL typesetting
\usepackage{booktabs}       % Professional-quality tables
\usepackage{amsfonts}       % Blackboard math symbols
\usepackage{nicefrac}       % Compact symbols for 1/2, etc.
\usepackage{tabularx}       % Tables
\usepackage{makecell}       % More for tables
\usepackage{ragged2e}       % Justification
\usepackage{paralist}       % Compact lists

\usepackage{import}         % Importing subdocuments
\usepackage{standalone}     % Compilable subdocuments and inclusion thereof
\usepackage{titlesec}       % Tweak titles

\usepackage{amsmath}        % Math
\usepackage{cases}          % Cases in math
\usepackage{amsthm}         % Theorems
\usepackage{thmtools}       % Tools for theorems
\usepackage{mathtools}      % More math commands
\usepackage{subcaption}     % Subcaptions
\usepackage{tikz}           % Figures
\usepackage{siunitx}        % SI units
\usepackage[
    noabbrev,
    capitalize,
    nameinlink]{cleveref}   % Automatic referencing

\begingroup
    \makeatletter
    \@for\theoremstyle:=theorem,corollary,lemma,model,definition,remark,plain\do{%
        \expandafter\g@addto@macro\csname th@\theoremstyle\endcsname{%
            \setlength\thm@preskip\parskip
            \setlength\thm@postskip\parskip%{0pt}
            \addtolength\thm@preskip\parskip
            }%
        }
    \makeatother
\endgroup

\theoremstyle{definition}
\newtheorem{theorem}{Theorem}
\newtheorem{lemma}{Lemma}
\newtheorem{definition}{Definition}
\newtheorem{proposition}{Proposition}

\crefname{theorem}{Theorem}{Theorems}
\Crefname{theorem}{Theorem}{Theorems}
\crefname{lemma}{Lemma}{Lemmas}
\Crefname{lemma}{Lemma}{Lemmas}
\crefname{proposition}{Proposition}{Propositions}
\Crefname{proposition}{Proposition}{Propositions}
\crefname{definition}{Definition}{Definitions}
\Crefname{definition}{Definition}{Definitions}

\newcolumntype{L}{>{\RaggedRight\arraybackslash}X}

\renewcommand{\vec}[1]{\boldsymbol{\mathrm{#1}}} % Vector
\newcommand{\mat}[1]{\vec{#1}}                   % Matrix
\renewcommand{\vec}[1]{#1}

\newcommand{\R}{\mathbb{R}}   % Real numbers
\renewcommand{\th}{\theta}     % Theta
\renewcommand{\O}{\mathcal{O}} % Landau's symbol

\newcommand{\diag}{\operatorname{diag}}

\newcommand{\Top}{\operatorname{T}}

\newcommand{\sd}[1]{\, \mathrm{d} #1}     % Straight 'd'
\newcommand{\cond}{\, | \,}               % Conditioning
\newcommand{\set}[1]{\{#1\}}              % Set
\newcommand{\vardot}{\,\cdot\,}           % Dot for a variable
\newcommand{\T}{\text{\textsf{T}}}        % Transpose
\newcommand{\iidsim}{\overset{\raisebox{-2pt}{\smash{\text{\tiny i.i.d.}}}}{\sim}}  % Compact \sim with iid above it
\newcommand{\rlh}{\rightleftharpoons}     % Two-sided harpoons

\usetikzlibrary{arrows.meta, fit}
\tikzset{
    x = 4.5em,
    y = 4em,
    latent/.style = {
        draw,
        circle,
        minimum size = 2.25em,
        inner sep = 0pt
    },
    arrow/.style = {
        -{Triangle[angle=45:.5em]}
    },
    plate/.style = {
        draw,
        shape = rectangle,
        inner sep = 5pt,
        rounded corners = 2pt
    }
}

\definecolor{tabgreen}{HTML}{59a14f}

\titlespacing{\section}{0pt}{.5\parskip}{.25\parskip}

\begin{document}

\twocolumn[
    \aistatstitle{The Gaussian Process Autoregressive Regression Model (GPAR)}
    \aistatsauthor{
        James Requeima$^{12}$
        \And
        Will Tebbutt$^{12}$
        \And
        Wessel Bruinsma$^{12}$
        \And
        Richard E. Turner$^{1}$
    }
    \runningauthor{James Requeima, Will Tebbutt, Wessel Bruisma, and Richard E. Turner}
    \aistatsauthor{
        \normalfont $^{1}$University of Cambridge
        and
        $^{2}$Invenia Labs, Cambridge, UK
    }
    \aistatsaddress{\texttt{\{jrr41,$\,$wct23,$\,$wpb23,$\,$ret26\}@cam.ac.uk}}
    \vspace{-.25cm}
]

\begin{abstract}
    \vspace{-.25cm}
    Multi-output regression models must exploit dependencies between outputs to maximise predictive performance. The application of Gaussian processes (GPs) to this setting typically yields models that are computationally demanding and have limited representational power. We present the Gaussian Process Autoregressive Regression (GPAR) model, a scalable multi-output GP model that is able to capture nonlinear, possibly input-varying, dependencies between outputs in a simple and tractable way:
    the product rule is used to decompose the joint distribution over the outputs into a set of conditionals, each of which is modelled by a standard GP. GPAR's efficacy is demonstrated on a variety of synthetic and real-world problems, outperforming existing GP models and achieving state-of-the-art performance on established benchmarks.
    \vspace{-.25cm}
\end{abstract}

\section{Introduction}
\label{sec:intro}
The Gaussian process (GP) probabilistic modelling framework provides a powerful and popular approach to nonlinear single-output regression \citep{Rasmussen:2006:Gaussian_Processes}. The popularity of GP methods stems from their modularity, tractability, and interpretability: it is simple to construct rich, nonlinear models by compositional covariance function design, which can then be evaluated in a principled way (e.g.\ via the marginal likelihood), before being interpreted in terms of their component parts. This leads to an attractive plug-and-play approach to modelling and understanding data, which is so robust that it can even be automated \citep{DuvLloGroetal13,Sun:2018:Differentiable_Compositional_Kernel_Learning_for}.

Most regression problems, however, involve multiple outputs rather than a single one.
When modelling such data, it is key to capture the dependencies between these outputs.
For example, noise in the output space might be correlated, or, whilst one output might depend on the inputs in a complex (deterministic) way, it may depend quite simply on other output variables. In both cases multi-output GP models are required. There is a plethora of existing multi-output GP models that can capture linear correlations between output variables if these correlations are fixed across the input space \citep{Goovaerts:1997:Geostatistics_for_Natural_Resources_Evaluation,Wackernagel:2003:Multivariate_Geostatistics,Teh:2005:Semiparametric_Latent_Factor,Bonilla:2008:Multi-Task_Gaussian_Process,Nguyen:2014:Collaborative_Multi-Output,Dai:2017:Efficient_Modeling_of_Latent_Information}.
However, one of the main reasons for the popularity of the GP approach is that a suite of different types of nonlinear \textit{input} dependencies can be modelled, and it is disappointing that this flexibility is not extended to interactions between the \textit{outputs}.
There are some approaches that do allow limited modelling of nonlinear output dependencies \citep{Wilson:2012:GP_Regression_Networks,Bruinsma:2016:GGPCM} but this flexibility comes from sacrificing tractability, with complex and computationally demanding approximate inference and learning schemes now required. This complexity significantly slows down model fitting, evaluation, and improvement work flow.

What is needed is a flexible and analytically tractable modelling approach to multi-output regression that supports plug-and-play modelling and model interpretation. The Gaussian Process Autoregressive Regression (GPAR) model, introduced in \cref{sec:gpar}, achieves these aims by taking an approach analogous to that employed by the Neural Autoregressive Density Estimator \citep{larochelle2011neural} for density modelling. The product rule is used to decompose the distribution of the outputs given the inputs into a set of one-dimensional conditional distributions. Critically, these distributions can be interpreted as a decoupled set of single-output regression problems, and learning and inference in GPAR therefore amount to a set of standard single-output GP regression tasks: training is closed form, fast, and amenable to standard scaling techniques.
GPAR converts the modelling of output dependencies that are possibly nonlinear and input-dependent into a set of standard GP covariance function design problems,
constructing expressive, jointly non-Gaussian models over the outputs.
Importantly, we show how GPAR can capture nonlinear relationships between outputs as well as structured, input-dependent noise, simply through kernel hyperparameter learning. We apply GPAR to a wide variety of multi-output regression problems, achieving state-of-the-art results on five benchmark tasks.

\section{GPAR}
\label{sec:gpar}
\begin{figure}
    \centering
    \begin{tikzpicture}[scale=.4]
        \begin{scope}[shift={(0, 0)}]
            \draw [thick, red, domain=0:3, smooth, variable=\x, samples=200] plot ({\x}, {1+.2*(\x-2.5)+.075*\x*\x+.3*sin(deg(1.5*2*pi*\x))});
            \draw [thick] (0, 2) -- (0, 0) -- (3, 0);
            \node [anchor=south] () at (1.5, 2) {CO${}_\text{2}$ $C(t)$};
            \node [anchor=north] (f1from1) at (1.5, 0) {};
            \node [anchor=west] (f1from2) at (3, 1.5) {};
        \end{scope}
        \begin{scope}[shift={(-2, -3.5)}]
            \node [anchor=west] () at (3, 1) {$=f_2(\;\;,\;\;,t)$};
            \node [anchor=center] (f3to1) at (4.7, .9) {};
            \node [anchor=center] (f3to2) at (5.2, .9) {};
            \draw [thick, red, domain=0:3, smooth, variable=\x, samples=200] plot ({\x}, {1-.125*(\x-1.5)+.6*sin(deg(1.5*2*pi*\x))});
            \draw [thick] (0, 2) -- (0, 0) -- (3, 0);
            \node [anchor=south] () at (1.5, 2) {Sea Ice $I(t)$};
        \end{scope}
        \begin{scope}[shift={(4.2, -2)}]
            \node [anchor=south, align=center] () at (1.5, 2.7) {$f_1(\;\;,t)$\\$=$};
            \node [anchor=center] (f2to1) at (1.6, 3.7) {};
            \draw [thick, red, domain=0:3, smooth, variable=\x, samples=200] plot ({\x}, {1+.125*(\x-1.5)+.6*sin(deg(1.5*2*pi*\x))});
            \draw [thick] (0, 2) -- (0, 0) -- (3, 0);
            \node [anchor=south] () at (1.5, 2) {Temperature $T(t)$};
            \node [anchor=east] (f2from1) at (0, 1) {};
        \end{scope}
        \draw [thick, ->, >={Triangle}] (f1from1) edge[out=-90, in=90] (f3to1);
        \draw [thick, ->, >={Triangle}] (f2from1) edge[out=180, in=90] (f3to2);
        \draw [thick, ->, >={Triangle}] (f1from2) edge[out=20, in=90] (f2to1);
    \end{tikzpicture}
    \caption{Cartoon motivating a factorisation for the joint distribution $p(I(t),T(t),C(t))$}
    \label{fig:cartoon}
    \vspace{-.25cm}
\end{figure}
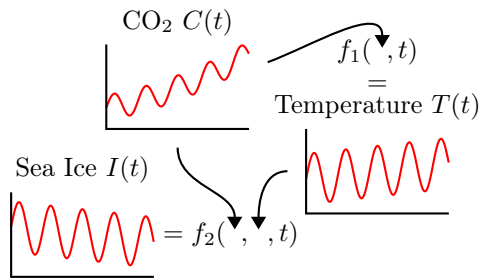

\begin{figure}
    \begin{subfigure}[t]{.48\linewidth}
        \centering
        \caption{}
        \begin{tikzpicture}[scale=.8]
            \node [latent] (f1) at (0, 0) {$f_1$};
            \node [latent] (f2) at (1, 0) {$f_2$};
            \node [latent] (f3) at (2, 0) {$f_3$};
            \node [latent] (y1) at (0, -1) {$y_1$};
            \node [latent] (y2) at (1, -1) {$y_2$};
            \node [latent] (y3) at (2, -1) {$y_3$};
            \node [plate, draw=white, fit=(y1)] () {};
            \draw [arrow] (f1) -- (y1);
            \draw [arrow] (f2) -- (y2);
            \draw [arrow] (f3) -- (y3);
            \draw [arrow] (y1) -- (y2);
            \draw [arrow] (y1) edge[bend right=40] (y3);
            \draw [arrow] (y2) -- (y3);
        \end{tikzpicture}
        \label{fig:graphical_model_nonlocal}
    \end{subfigure}
    \hfill
    \begin{subfigure}[t]{.48\linewidth}
        \centering
        \caption{}
        \begin{tikzpicture}[scale=.8]
            \node [latent] (f1) at (-1, 0) {$f_1$};
            \node [latent] (f2) at (0, 0) {$f_2$};
            \node [latent] (f3) at (1, 0) {$f_3$};
            \node [latent] (y1x) at (-1, -1) {$y_1(x)$};
            \node [latent] (y2x) at (0, -1) {$y_2(x)$};
            \node [latent] (y3x) at (1, -1) {$y_3(x)$};
            \draw [arrow] (f1) -- (y1x);
            \draw [arrow] (f2) -- (y2x);
            \draw [arrow] (f3) -- (y3x);
            \draw [arrow] (y1x) -- (y2x);
            \draw [arrow] (y1x) edge [bend right=40] (y3x);
            \draw [arrow] (y2x) -- (y3x);
            \node [plate,
                   inner ysep=8pt,
                   yshift=-6pt,
                   label={[anchor=south east]south east:$x$},
                   fit=(y1x) (y2x) (y3x)] () {};
        \end{tikzpicture}
        \label{fig:graphical_model}
    \end{subfigure}
    \caption{Graphical models corresponding to (a) nonlocal GPAR and (b) GPAR}
    \vspace{-.25cm}
\end{figure}
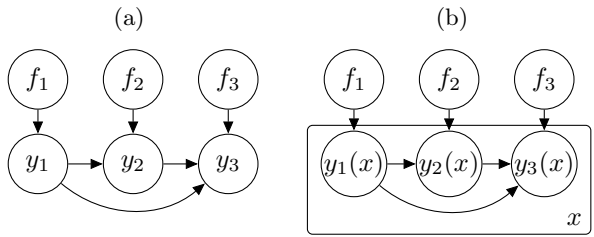

\newcommand{\cotwo}{CO${}_\text{2}$}
Consider the problem of modelling the world's average \cotwo~level $C(t)$, temperature $T(t)$, and Arctic sea ice extent $I(t)$ as a function of time $t$.
By the greenhouse effect, one can imagine that the temperature $T(t)$ is a complicated, stochastic function $f_1$ of \cotwo~and time: $T(t)=f_1(C(t),t)$.
Similarly, one might hypothesise that the Arctic sea ice extent $I(t)$ can be modelled as another complicated function $f_2$ of temperature, \cotwo, and time: $I(t)=f_2(T(t),C(t),t)$.
These functional relationships are depicted in \cref{fig:cartoon} and motivate the following model where the conditionals model the postulated underlying functions $f_1$ and $f_2$:
\begin{align*}
    &p(I(t),T(t),C(t))\\
    &=p(C(t)) \underbrace{p(T(t)\cond C(t))}_{\text{models $f_1$}} \underbrace{p(I(t)\cond T(t), C(t))}_{\text{models $f_2$}} .
\end{align*}
More generally, consider the problem of modelling $M
$ outputs $y_{1:M}(x)=(y_1(x),\ldots,y_M(x))$ as a function of the input $x$.
Applying the product rule yields\footnote{This requires an ordering of the outputs; we will address this point in \cref{sec:gpar}.}
\begin{align}
    &p(y_{1:M}(x)) \label{eq:product_rule} \\
    &=p(y_1(x))
    \underbrace{p(y_2(x)\cond y_1(x))}_{\substack{\text{$y_2(x)$ as a random} \\ \text{function of $y_1(x)$}}}
    \cdots\;
    \underbrace{p(y_M(x)\cond y_{1:M-1}(x)),}_{\substack{\text{$y_M(x)$ as a random}\\ \text{function of $y_{1:M-1}(x)$}}} \nonumber
\end{align}
which states that $y_1(x)$, like \cotwo, is first generated from $x$, according to some unknown, random function $f_1$; that $y_2(x)$, like temperature, is then generated from $y_1(x)$ and $x$, according to some unknown, random function $f_2$; that $y_3(x)$, like the Arctic sea ice extent, is then generated from $y_2(x)$, $y_1(x)$, and $x$, according to some unknown, random function $f_3$; {\it et cetera}:
\begin{align*}
    y_1(x) &= f_1(x), & f_1 &\sim p(f_1), \\
    y_2(x) &= f_2(y_1(x), x), & f_2 &\sim p(f_2), \\
     &\,\;\vdots \\
    y_M(x) &= f_M(y_{1:M-1}(x),x) & f_M & \sim p(f_M).
\end{align*}
GPAR, introduced now, models these unknown functions $f_{1:M}$ with Gaussian processes (GPs).
Recall that a GP $f$ over an index set $\mathcal{X}$ defines a stochastic process, or process in short, where $f(x_1),\ldots,f(x_N)$ are jointly Gaussian distributed for any $x_1,\ldots,x_N$ \citep{Rasmussen:2006:Gaussian_Processes}.
Marginalising out $f_{1:M}$, we find that GPAR models the conditionals in \cref{eq:product_rule} with Gaussian processes:
\begin{align}
    &y_m\cond y_{1:m-1} \label{eq:gpar_conditionals} \\
    &\quad\sim \mathcal{GP}(0, k_m((y_{1:m-1}(x),x), (y_{1:m-1}(x'),x'))). \nonumber
\end{align}
Although the conditionals in \cref{eq:product_rule} are Gaussian, the joint distribution $p(y_{1:N})$ is not;
moments of the joint distribution over the outputs are generally intractable, but samples can be generated by sequentially sampling the conditionals.
\Cref{fig:graphical_model} depicts the graphical model corresponding to GPAR.
Crucially, the kernels $(k_m)$ may specify nonlinear, input-dependent relationships between outputs, which enables GPAR to model data where outputs inter-depend in complicated ways.

Returning to the climate modelling example, one might object that the temperature $T(t)$ at time $t$ does not just depend the \cotwo~level $C(t)$ at time $t$, but instead depends on the entire history of \cotwo~levels $C$: $T(t) = f_1(C,t)$.
\textit{Note: by writing $C$ instead of $C(t)$, we refer to the entire function $C$, as opposed to just its value at $t$.}
Similarly, one might object that the Arctic sea ice extent $I(t)$ at time $t$ does not just depend on the temperature $T(t)$ and \cotwo~level $C(t)$ at time $t$, but instead depends on the entire history of temperatures $T$ and \cotwo~levels $C$: $T(t) = f_2(T,C,t)$.
This kind of dependency structure, where output $y_m(x)$ now depends on the entirety of all foregoing outputs $y_{1:m}$ instead of just their value at $x$, motivates the following generalisation of GPAR in its form of \cref{eq:gpar_conditionals}:
\begin{align}\label{eq:gpar_nonlocal_conditionals}
    y_m\cond y_{1:m\!-\!1} 
    &\!\sim\! \mathcal{GP}(0, k_m((y_{1:m\!-\!1},x), (y_{1:m\!-\!1},x'))),
\end{align}
which we refer to as \textit{nonlocal} GPAR, or \textit{non-instantaneous} in the temporal setting.
Clearly, GPAR is a special case of nonlocal GPAR.
\Cref{fig:graphical_model_nonlocal} depicts the graphical model corresponding to nonlocal GPAR.
Nonlocal GPAR will not be evaluated experimentally, but some theoretical properties will be described.

\textbf{Inference and learning in GPAR}. Inference and learning in GPAR is simple.
Let $y^{(n)}_m=y_m(x^{(n)})$ denote an observation for output $m$.
Then, assuming all outputs are observed at each input, we find
\begin{align} \label{eq:posterior}
    &p(f_{1:M}\cond (y^{(n)}_{1:M}, x^{(n)})_{n=1}^N) \\
    &= \prod_{m=1}^M p(f_m\cond \underbrace{(y^{(n)}_m)_{n=1}^N}_{\substack{\text{observations }\\\text{for $f_m$}}},\,\underbrace{(y^{(n)}_{1:m-1}, x^{(n)})_{n=1}^N}_{\substack{\text{input locations}\\\text{of observations}}}\;), \nonumber
\end{align}
meaning that the posterior of $f_m$ is computed simply by conditioning $f_m$ on the observations for output $m$ located at the observations for the foregoing outputs and the input, which again is a Gaussian process;
thus, like the prior, the posterior predictive density also decomposes as a product of Gaussian conditionals.
The evidence $\log p((y_{1:M}^{(n)})_{n=1}^N\cond(x^{(n)})_{n=1}^N)$ decomposes similarly; as a consequence, the hyperparameters for $f_m$ can be learned simply by maximising $\log p((y^{(n)}_m)_{n=1}^N\cond (y^{(n)}_{1:m-1}, x^{(n)})_{n=1}^N)$.
In conclusion, \textit{inference and learning in GPAR with $M$ outputs comes down to inference and learning in $M$ decoupled, one-dimensional GP regression problems}.
This shows that without approximations GPAR scales $\O(M N^3)$, depending only linearly on the number of outputs $M$ rather than cubically, as is the case for general multi-output GPs, which scale $\O(M^3 N^3)$.

\textbf{Scaling GPAR}. In these one-dimensional GP regression problems, off-the-shelf GP scaling techniques may be applied to trivially scale GPAR to large data sets.
Later we utilise the variational inducing point method by \citet{Titsias:2009:Variational_Learning}, which is theoretically principled \citep{Matthews:2016:On_Sparse_Variational} and empirically accurate \citep{Bui:2016:A_Unifying_Framework_for_Gaussian}.
This method requires a set of inducing inputs to be specified for each Gaussian process. For the first output $y_1$ there is a single input $x$, which is time. We use fixed (non-optimised) regularly spaced inducing inputs, as they are known to perform well for time series \citep{Bui:2014:Tree-Structured_Gaussian}. 
The second and following outputs $y_m$, however, require inducing points to be placed in $x$ \emph{and} $y_1,\ldots,y_{m-1}$. Regular spacing can be used again for $x$, but there are choices available for $y_1,\ldots,y_{m-1}$. One approach would be to optimise these inducing input locations, but instead
we use the posterior predictive means of $y_1,\ldots,y_{m-1}$ at $x$. This choice accelerates training and was found to yield good results.

\textbf{Missing data}. When there are missing outputs, the procedures for inference and learning described above remain valid as long as for every observation $y_m^{(n)}$ there are also observations $y_{m-1:1}^{(n)}$.
We call a data set satisfying this property \textit{closed downwards}.
If a data set is not closed downwards, data may be imputed, e.g.\ using the model's posterior predictive mean, to ensure closed downwardness; inference and learning then, however, become approximate. The key conditional independence expressed by \cref{fig:graphical_model} that results in simple inference and learning is the following:
\begin{theorem} \label{thm:d-separation-in-text}
    Let a set of observations $\mathcal{D}$ be closed downwards.
    Then $y_{i} \perp \mathcal{D}_{i+1:M} \cond \mathcal{D}_{1:i}$, where $\mathcal{D}_{1:i}$ are the observations for outputs $1,\ldots,i$ and $\mathcal{D}_{i+1:M}$ for $i+1,\ldots,M$.
    % \footnote{
    %     $\mathcal{D}_{1:i}=\set{y_j^{(n)} \in \mathcal{D}: j \le i,\, n \le N}$ and 
    %     $\mathcal{D}_{i+1:M}=\set{y_j^{(n)} \in \mathcal{D}: j > i,\, n \le N}$.
    % }
\end{theorem}
\Cref{thm:d-separation-in-text} is proved in Appendix A from the supplementary material.
Note that \cref{thm:d-separation-in-text} holds for every graphical model of the form of \cref{fig:graphical_model}, meaning that the decomposition into single-output modelling problems holds for \textit{any} choice of the conditionals in \cref{eq:product_rule}, not just Gaussians.

\textbf{Potential deficiencies of GPAR}. GPAR has two apparent limitations. First, since the outputs from earlier dimensions are used as inputs to later dimensions, noisy outputs yield noisy inputs.
One possible mitigating solution is to employ a \textit{denoising} input transformation for the kernels, e.g.\ using the posterior predictive mean of the foregoing outputs as the input of the next covariance function. 
We shall refer to GPAR models employing this approach as D-GPAR.
Second, one needs to choose an ordering of the outputs.
Fortunately, often the data admits a natural ordering; for example, if the predictions concern a single output dimension, this should be placed last.
Alternatively, if there is not a natural ordering, one can greedily optimise the evidence with respect to the ordering.
This procedure considers $\tfrac{1}{2}M(M+1)$ configurations while an exhaustive search would consider all $M!$ possible configurations:
the best first output is selected out of all $M$ choices, which is then fixed;
then, the best second output is selected out of the remaining $M-1$ choices, which is then fixed; et cetera.
These methods are examined in \cref{sec:experiments}.

\section{GPAR and Multi-Output GPs}

The choice of kernels $k_{1:M}$ for $f_{1:M}$ is crucial to GPAR, as they determine the types of relationships between inputs and outputs that can be learned.
Particular choices for $k_{1:M}$ turn out to yield models closely related to existing work.
These connections are made rigorous by the nonlinear and linear equivalent model discussed in Appendix B from the supplementary material.
We summarise the results here; see also \Cref{tab:connections}.

If $k_{m}$ depends linearly on the foregoing outputs $y_{1:m-1}$ at particular $x$, then a joint Gaussian distribution over the outputs is induced in the form of a multi-output GP model \citep{Goovaerts:1997:Geostatistics_for_Natural_Resources_Evaluation,Stein:1999:Interpolation_of_Spatial_Data,Wackernagel:2003:Multivariate_Geostatistics,Teh:2005:Semiparametric_Latent_Factor,Bonilla:2008:Multi-Task_Gaussian_Process,Nguyen:2014:Collaborative_Multi-Output} where latent processes are mixed together according to a matrix, called the \textit{mixing matrix}, that is \textit{lower-triangular} (Appendix B).
One may let the dependency of $k_m$ on $y_{1:m-1}$ vary with $x$, in which case the mixing matrix varies with $x$, meaning that correlations between outputs vary with $x$.
This yields an instance of the Gaussian Process Regression Network (GPRN) \citep{Wilson:2012:GP_Regression_Networks} where inference is fast and closed form.
One may even let $k_m$ depend nonlinearly on $y_{1:m-1}$, which yields a particularly structured deep Gaussian process (DGP) \citep{Damianou:2015:Deep_Gaussian_Processes_and_Variational,Bui:2016:Deep_Gaussian_Processes_for_Regression}, potentially with skip connections from the inputs (Appendix B).
Note that GPAR may be interpreted as a conventional DGP where the hidden layers are directly observed and correspond to successive outputs; this connection could potentially be leveraged to bring machinery developed for DGPs to GPAR, e.g.\ to deal with arbitrarily missing data.

One can further let $k_{m}$ depend on the \textit{entirety} of the foregoing outputs $y_{1:m-1}$, yielding instances of nonlocal GPAR. 
An example of a nonlocal linear kernel is
\begin{align*}
    k((y, x), (y', x')) = \!\int\! a(x - z, x' - z')y(z)y'(z')\sd{z}\sd{z'}.
\end{align*}
The nonlocal linear kernel again induces a jointly Gaussian distribution over the outputs in the form of a convolutional multi-output GP model \citep{Alvarez:2009:Latent_Force_Models,Alvarez:2009:Sparse_Convolved_Gaussian_Processes_for,Bruinsma:2016:GGPCM}
where latent processes are convolved together according to a matrix-valued function, called the \textit{convolution matrix}, that is \textit{lower-triangular} (Appendix B).
Again, one may let the dependency of $k_m$ on the entirety of $y_{1:m-1}$ vary with $x$, in which case the convolution matrix varies with $x$,
or even let $k_m$ depend nonlinearly on the entirety of $y_{1:m-1}$;
an example of a nonlocal nonlinear kernel is
\begin{align*}
    k(y, y') = \sigma^2 \exp\left({-\!\!\int}\frac{1}{2\ell(z)}(y(z) - y'(z))^2\sd{z}\right).
\end{align*}
Henceforth, we shall refer to GPAR with linear dependencies between outputs as GPAR-L, GPAR with nonlinear dependencies between outputs as GPAR-NL, and a combination of the two as GPAR-L-NL.

\begin{table*}[htbp]
    \centering
    \centerline{\begin{tabular}{rllp{4.75cm}}
        \toprule
        GPAR & \makecell[vl]{Deps. Between Outputs} & Kernels for $f_{1:M}$ & Related Models \\ \midrule
        Local
            & Linear
                & $k_1(x, x') + k_{\text{Linear}}(y(x), y(x'))$
                & Multi-Output GPs [1] \\
            & \hspace{.5em} $+$ dep.\ on $x$
                & $k_1(x, x') + k_2(x, x') k_{\text{Linear}}(y(x), y(x'))$
                & GPRN [2] \\
            & Nonlinear
                & $k_1(x, x') + k_2(y(x), y(x'))$
                & Deep GPs (DGPs) [3]  \\
            & \hspace{.5em} $+$ dep.\ on $x$
                & $k_1(x, x') + k_2((x, y(x)), (x', y(x')))$
                & DGPs with input connections \\
        \midrule
        Nonlocal
             & Linear
                & $k_1(x, x') + k_{\text{Linear}}(y, y')$
                & \\
            & \hspace{.5em} $+$ dep.\ on $x$
                & $k_1(x, x') + k_2(x, x')k_{\text{Linear}}(y, y')$
                & Convolutional MOGPs [4] \\
            & Nonlinear
                & $k_1(x, x') + k_2(y, y')$
                & \\
            & \hspace{.5em} $+$ dep.\ on $x$
                & $k_1(x, x') + k_2((x, y), (x', y'))$
                & \\
        \bottomrule
    \end{tabular}}
    \caption{
        Classification of kernels $k_{1:M}$ for $f_{1:M}$, the resulting dependencies between outputs, and related models.
        Here $k_{\text{Linear}}$ refers to a linear kernel and $k_1$ and $k_2$ to an exponentiated quadratic (EQ) or rational quadratic (RQ) kernel \citep{Rasmussen:2006:Gaussian_Processes}.
        {%\small
            [1]: \citet{Goovaerts:1997:Geostatistics_for_Natural_Resources_Evaluation,Stein:1999:Interpolation_of_Spatial_Data,Wackernagel:2003:Multivariate_Geostatistics,Teh:2005:Semiparametric_Latent_Factor,Bonilla:2008:Multi-Task_Gaussian_Process,Osborne:2008:Towards_Real-Time_Information_Processing_of,Nguyen:2014:Collaborative_Multi-Output}.
            [2]: \citet{Wilson:2012:GP_Regression_Networks}.
            [3]: \citet{Damianou:2015:Deep_Gaussian_Processes_and_Variational,Bui:2016:Deep_Gaussian_Processes_for_Regression}.
            [4]: \citet{Alvarez:2009:Latent_Force_Models,Alvarez:2009:Sparse_Convolved_Gaussian_Processes_for,Bruinsma:2016:GGPCM}.
        }
    }
    \label{tab:connections}
    \vspace{-.25cm}
\end{table*}

\section{Further Related Work}
\label{sec:related}
The Gaussian Process Network \citep{friedman2000gaussian} is similar to GPAR, but was instead developed to identify causal dependencies between variables in a probabilistic graphical models context rather than multi-output regression. The work by \citet{yuan2011conditional} also discusses a model similar to GPAR, but specifies a different generative procedure for the outputs.

The multi-fidelity modelling literature is closely related to multi-output modelling. Whereas in a multi-output regression task we predict all outputs,  multi-fidelity modelling is concerned with predicting a particular high-fidelity function, incorporating information from observations from various levels of fidelity. The idea of iteratively conditioning on lower fidelity models in the construction of higher fidelity ones is a well-used strategy \citep{kennedy2000predicting, le2014recursive}. The model presented by \citet{perdikaris2017nonlinear} is nearly identical to GPAR applied in the multi-fidelity framework, but applications outside this setting have not been considered.

Moreover, GPAR follows a long line of work on the family of fully visible
Bayesian networks \citep{frey1996does,bengio2000modeling} that decompose the distribution over the observations according to the product rule (\cref{eq:product_rule}) and model the resulting one dimensional conditionals.
A number of approaches use neural networks for this purpose \citep{neal1992connectionist,frey1996does,larochelle2011neural,theis2015generative,van2016conditional}. 
In particular, if the observations are real-valued, a standard architecture lets the conditionals be Gaussian with means encoded by neural networks and fixed variances.
Under broad conditions, if these neural networks are replaced by Bayesian neural networks with independent Gaussian priors over the weights, we recover GPAR as the width of the hidden layers goes to infinity \citep{neal1996bayesian,matthews2018gaussian}.

\section{Synthetic Data Experiments}
\label{sec:synthetic}
GPAR is well suited for problems where there is a strong functional relationship between outputs and for problems where observation noise is richly structured and input dependent. In this section we demonstrate GPAR's ability to model both types of phenomena.

\begin{figure*}[t]
    \centering
    \includegraphics[width=\linewidth]{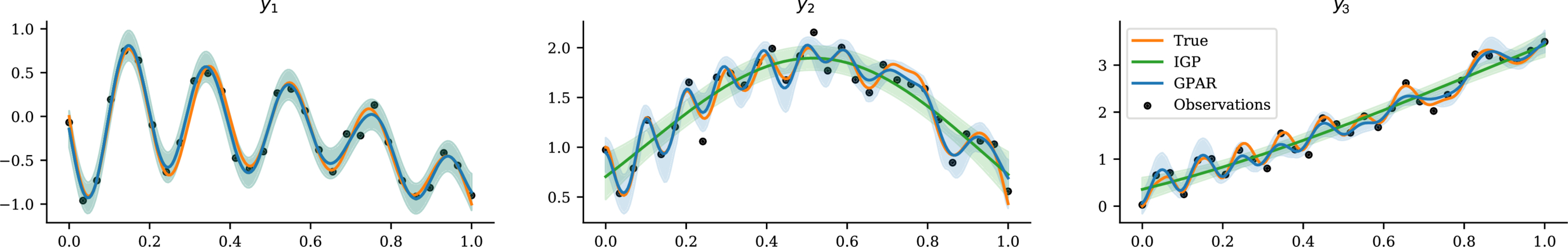}
    \vspace{-1em}
    \caption{
        Synthetic data set with complex output dependencies: GPAR vs independent GPs (IGP) predictions.
    }
    \label{fig:functional_dependency}
    \vspace{-.2cm}
\end{figure*}

First, we test the ability to leverage strong functional relationships between the outputs.
Consider three outputs $y_1$, $y_2$, and $y_3$, inter-depending nonlinearly:
\begin{align*}
    y_1(x) &= -\sin(10\pi(x+1))/(2x + 1) - x^4 + \epsilon_1, \\
    y_2(x) &= \cos^2(y_1(x)) + \sin(3x) + \epsilon_2, \\
    y_3(x) &= y_2(x)y_1^2(x) + 3x + \epsilon_3,
\end{align*}
where $\epsilon_1, \epsilon_2, \epsilon_3 \iidsim \mathcal{N}(0, 0.05)$.
By substituting $y_1$ and $y_2$ into $y_3$, we see that $y_3$ can be expressed directly in terms of $x$, but via a complex function. The dependence of $y_3$ on $y_1$, $y_2$, however, is much simpler.
Therefore, as GPAR can exploit direct dependencies between $y_{1:2}$ and $y_3$, it should be presented with a much simplified task as compared to predicting $y_3$ from $x$ directly.
\Cref{fig:functional_dependency} shows plots of independent GPs (IGPs) and GPAR fit to $30$ data points from $y_1$, $y_2$ and $y_3$.
Indeed observe that GPAR is able to learn $y_2$'s dependence on $y_1$, and $y_3$'s dependence on $y_1$ and $y_2$, whereas the independent GPs struggle with the complicated structure.

\begin{figure}[t]
    \centering
    \includegraphics[width=\linewidth, trim={0 1.5cm 2.5cm 3cm}]{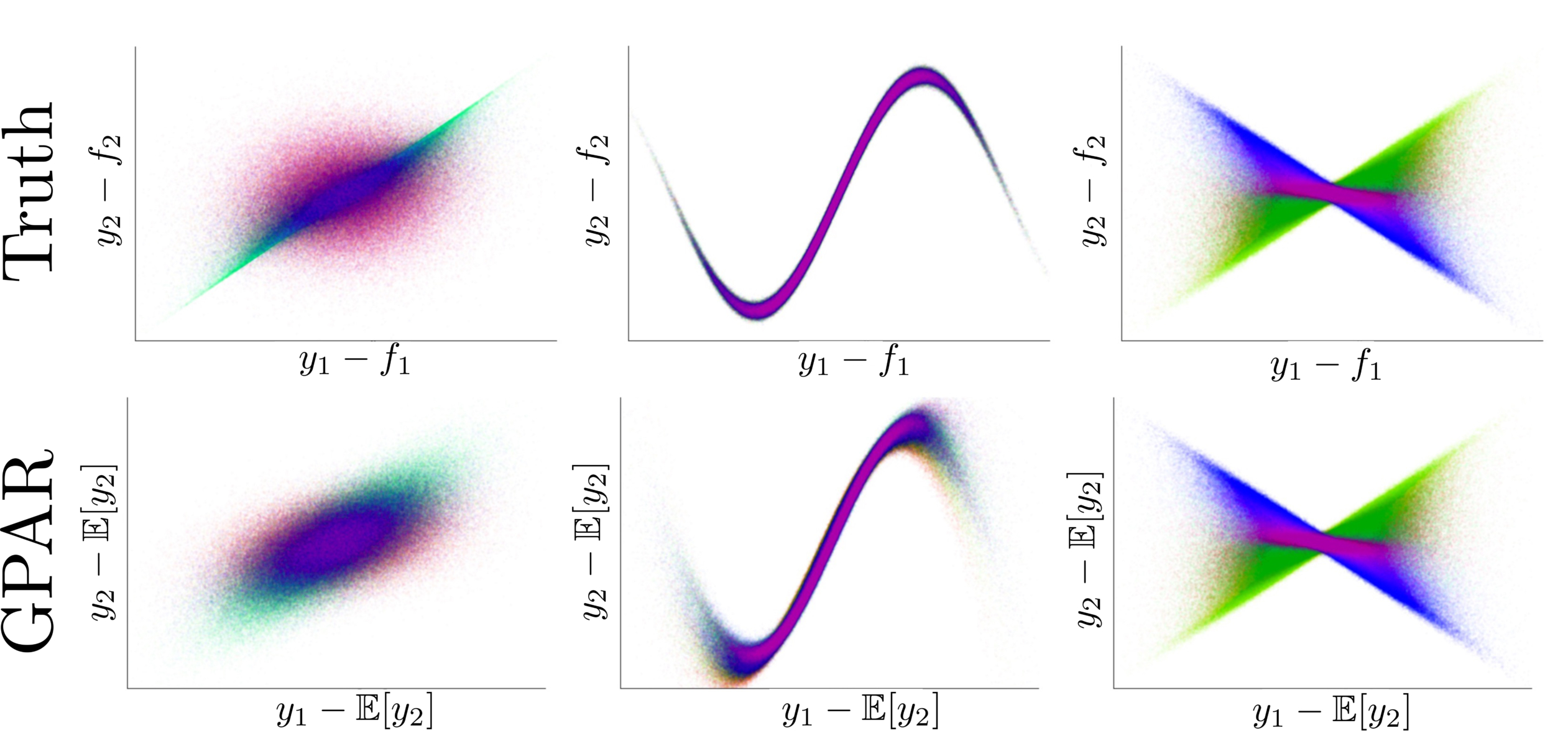}
    \caption{Correlation between the sample residues (deviation from the mean) for $y_1$ and $y_2$. Left, middle, and right plots correspond to schemes (1), (2) and (3) respectively. Samples are coloured according to input value $x$; that is, all samples for a particular $x$ have the same colour. If the colour pattern is preserved, then GPAR has successfully captured how the noise in $y_1$ correlates to that in $y_2$.}
    \label{fig:noise_plots}
\end{figure}

Second, we test GPAR's ability to capture non-Gaussian and input-dependent noise. Consider the following three schemes in which two outputs are observed under various noise conditions: $y_1(x) = f_1(x) + \epsilon_1$ and
\begin{align*}
    \text{(1): } y_2(x) &= f_2(x) + \sin^2(2 \pi x)\epsilon_1 + \cos^2(2 \pi x) \epsilon_2, \\
    \text{(2): } y_2(x) &= f_2(x) + \sin(\pi\epsilon_1) + \epsilon_2, \\
    \text{(3): } y_2(x) &= f_2(x) + \sin(\pi x) \epsilon_1 + \epsilon_2,
\end{align*}
where $\epsilon_1,\epsilon_2 \iidsim \mathcal{N}(0, 0.1)$, and $f_1$ and $f_2$ are complicated, nonlinear functions.\footnote{The functions given by $f_1(x) = -\sin(10\pi(x+1))/(2x + 1) - x^4$ and $f_2(x) = \frac15 e^{2x}\left(\th_{1}\cos(\th_{2} \pi x) + \th_{3}\cos(\th_{4} \pi x) \right) + \sqrt{2x}$.}
All three schemes have i.i.d.\ homoscedastic Gaussian noise in $y_1$.
The noise in $y_2$, however, depends on that in $y_1$ and can be heteroscadastic.
The task for GPAR is to learn the scheme's noise structure. \Cref{fig:noise_plots} visualises the noise correlations induced by the schemes and the noise structures learned by GPAR.
Observe that GPAR is able to learn the various noise structures.

\section{Real-World Data Experiments}
\label{sec:experiments}

\begin{table}[t]
    \setlength\tabcolsep{1.5pt}
    \centerline{\begin{tabular}{p{.2\linewidth}p{.8\linewidth}}
        \toprule
        Acronym&Model  \\ \midrule
        IGP & Independent GPs \\
        CK & Cokriging  \\
        ICM & Intrinstic Coregionalisation Model [1] \\
        SLFM & Semi-Parametric Latent Factor Model [2]  \\
        CGP & Collaborative Multi-Output GPs [3]\\
        CMOGP & Convolved Multi-Output GP Model [4]\\
        GPRN & GP Regression Network [5]\\
        \bottomrule
    \end{tabular}}
    \caption{
        Models against which GPAR is compared.
        {%\small
            [1]: \citet{Goovaerts:1997:Geostatistics_for_Natural_Resources_Evaluation,Stein:1999:Interpolation_of_Spatial_Data,Wackernagel:2003:Multivariate_Geostatistics}.
            [2]: \citet{Teh:2005:Semiparametric_Latent_Factor}.
            [3]: \citet{Nguyen:2014:Collaborative_Multi-Output}.
            [4]: \citet{Alvarez:2011:Computationally_Efficient_Convolved,Alvarez:2010:Efficient_Multioutput_Gaussian_Processes_Through}.
            [5]: \citet{Wilson:2012:GP_Regression_Networks}.
        }    
    }
    \label{tab:models_to_compare_against}
    \vspace{-.5cm}
\end{table}

In this section we evaluate GPAR's performance and compare to other models on four standard data sets commonly used to evaluate multi-output models.
We also consider a recently-introduced data set in the field of Bayesian optimisation, which is a downstream application area that could benefit from GPAR.
\Cref{tab:models_to_compare_against} lists the models against which we compare GPAR.
We always compare against IGP and CK, ICM, SLFM, and CGP, and compare against CMOGP and GPRN if results for the considered task are available.
Since CK and ICM are much simplified versions of SLFM \citep{Alvarez:2010:Efficient_Multioutput_Gaussian_Processes_Through,Goovaerts:1997:Geostatistics_for_Natural_Resources_Evaluation} and CGP is an approximation to SLFM, we sometimes omit results for CK, ICM, and CGP.
Implementations can be found at \url{https://github.com/wesselb/gpar} (Python) and \url{https://github.com/willtebbutt/GPAR.jl} (Julia).
Experimental details can be found in Appendix D from the supplementary material.

\textbf{Electroencephalogram (EEG) data set}.\footnote{
    The EEG data set can be downloaded at \url{https://archive.ics.uci.edu/ml/datasets/eeg+database}.
}
This data set consists of 256 voltage measurements from 7 electrodes placed on a subject's scalp whilst the subject is shown a certain image; \citet{Zhang:1995:Event_Related_Potentials_During_Object} describe the data collection process in detail. In particular, we use frontal electrodes FZ and F1--F6 from the first trial on control subject \num{337}. The task is to predict the last 100 samples for electrodes FZ, F1, and F2, given that the first 156 samples of FZ, F1, and F2 and the whole signals of F3--F6 are observed. Performance is measured with the standardised mean squared error (SMSE), mean log loss (MLL) \citep{Rasmussen:2006:Gaussian_Processes}, and training time (TT).
\begin{figure*}[t]
    \centering
    \centerline{\includegraphics[width=\linewidth, trim={0 .75cm 0 0}]{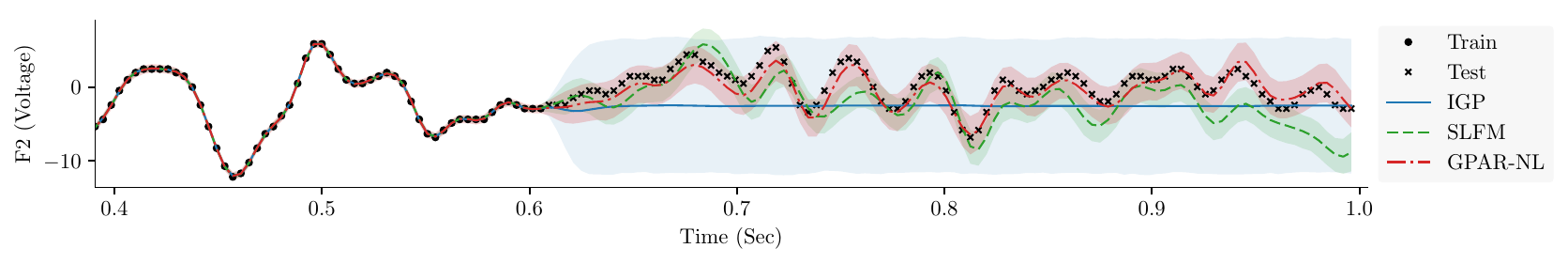}}
    \caption{Predictions for electrode F2 from the EEG data set}
    \label{fig:eeg_prediction}
    \vspace{-.25cm}
\end{figure*}
\begin{table}[t]
    \centering
    \begin{tabular}{lccc}
        \toprule
        Model & SMSE & MLL & TT \\ \midrule
        IGP & $1.75$ & $2.60$ & \SI{2}{sec} \\
        SLFM & $1.06$& $4.00$ & \SI{11}{min} \\
        GPAR-NL & $\mathbf{0.26}$ & $\mathbf{1.63}$ & \SI{5}{sec} \\
        \bottomrule
    \end{tabular}
    \caption{Results for the EEG data set for IGP, the SLFM with four latent dimensions, and GPAR.}
    \label{tab:eeg_results}
    \vspace{-.25cm}
\end{table}
\Cref{fig:eeg_prediction} visualises predictions for electrode F2, and \cref{tab:eeg_results} quantifies the results. We observe that GPAR-NL outperforms independent  in terms of SMSE and MLL; note that independent GPs completely fail to provide an informative prediction. Furthermore, independent GPs were trained in two seconds, and GPAR-NL took only three more seconds; in comparison, training SLFM took 11 minutes.

\textbf{Jura data set}.\footnote{
    The data can be downloaded at \url{https://sites.google.com/site/goovaertspierre/pierregoovaertswebsite/download/}.
}
This data set comprises metal concentration measurements collected from the topsoil in a \SI{14.5}{km^2} region of the Swiss Jura. We follow the experimental protocol by \citet{Goovaerts:1997:Geostatistics_for_Natural_Resources_Evaluation} also followed by \citet{Alvarez:2011:Computationally_Efficient_Convolved}: The training data comprises 259 data points distributed spatially with three output variables---nickel, zinc, and cadmium---and 100 additional data points for which only two of the three outputs---nickel and zinc---are observed. The task is to predict cadmium at the locations of those 100 additional data. Performance is evaluated with the mean absolute error (MAE). 

\Cref{tab:jura_results} shows the results.
The comparatively poor performance of independent GPs  highlights the importance of exploiting correlations between the mineral concentrations. Furthermore,
\cref{tab:jura_results} shows that D-GPAR-NL significantly outperforms the other models, achieving a new state-of-the-art.

\begin{table}[t]
    \centering
    \centerline{\begin{tabular}{lccccc}
        \toprule
        Model     & IGP      & CK${}^\dagger$ & ICM      & SLFM     & CMOGP${}^\dagger$ \\
        \midrule
        MAE       & $0.5739$ & $0.51$                & $0.4601$ & $0.4606$ & $0.4552$         \\
        MAE${}^*$ & $0.5753$ &                       & $0.4114$ & $0.4145$ &                  \\
        \bottomrule
    \end{tabular}}~\\[.1em]
    \begin{tabular}{lccc}
        \toprule
        Model     & GPRN${}^\dagger$ & GPAR-NL & D-GPAR-NL \\
        \midrule
        MAE       & $0.4525$         & $0.4324$ & $\mathbf{0.4114}$ \\
        MAE${}^*$ & $0.4040$         & $0.4168$ & $\mathbf{0.3996}$ \\
        \bottomrule
    \end{tabular}\hfill
    \caption{Results for the Jura data set for IGP, cokriging (CK) and ICM with two latent dimensions, the SLFM with two latent dimensions, CMOGP, GPRN, and GPAR. \quad ${}^*$  Results are obtained by first $\log$-transforming the data, then performing prediction, and finally transforming the predictions back to the original domain. \quad ${}^\dagger$ Results from \citet{Wilson:2014:Covariance_Kernels_for_Fast_Automatic}.}
    \label{tab:jura_results}
    \vspace{-.5cm}
\end{table}

\textbf{Exchange rates data set}.\footnote{
    The exchange rates data set can be downloaded at \url{http://fx.sauder.ubc.ca}.
}
This data set consists of the daily exchange rate w.r.t.\ USD of the top ten international currencies (CAD, EUR, JPY, GBP, CHF, AUD, HKD, NZD, KRW, and MXN) and three precious metals (gold, silver, and platinum) in the year 2007. The task is to predict CAD on days 50--100, JPY on days 100--150, and AUD on days 150--200, given that CAD is observed on days 1--49 and 101--251, JPY on days 1--99 and 151--251, and AUD on days 1--149 and 201--251; and that all other currencies are observed throughout the whole year.
Performance is measured with the SMSE.

\begin{figure*}[t]
    \centering
    \includegraphics[width=\linewidth, trim={0 0 5cm 0}]{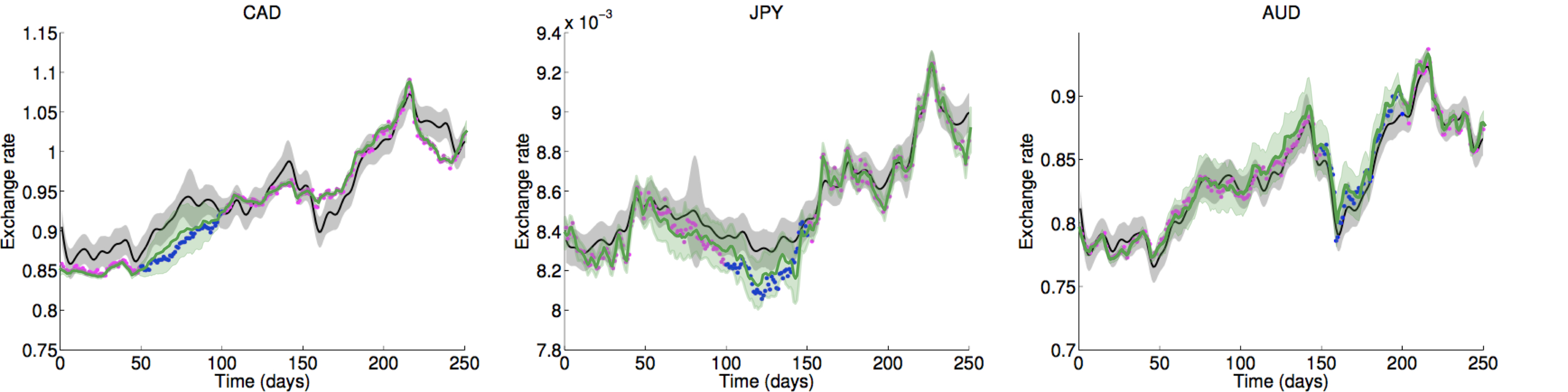}
    \caption{Visualisation of the exchange rates data set and CGP's (black) and GPAR's ({\color{tabgreen}green}) predictions for it. GPAR's predictions are overlayed on the original figure by \citet{Nguyen:2014:Collaborative_Multi-Output}.}
    \label{fig:ex_prediction}
\end{figure*}

\Cref{fig:ex_prediction} visualises GPAR's prediction for data set, and \cref{tab:ex_results} quantifies the result.
We greedily optimise the evidence w.r.t.\ the ordering of the outputs to determine an ordering, and we impute missing data to ensure closed downwardness of the data.
Observe that GPAR significantly outperforms all other models.

\begin{table}[t]
    \centerline{\begin{tabular}{lccccc}
        \toprule
        Model & IGP${}^*$ & CMOGP${}^*$ & CGP${}^*$ & GPAR-L-NL\\ \midrule
        SMSE  & $0.5996$  & $0.2427$    & $0.2125$ & $\mathbf{0.0302}$   \\
        \bottomrule
    \end{tabular}}
    % Without data imputation: 0.0374
    \caption{Experimental results for the exchange rates data set for IGP, CMOGP, CGP, and GPAR. \quad ${}^*$ These numbers are taken from \citet{Nguyen:2014:Collaborative_Multi-Output}.}
    \label{tab:ex_results}
    \vspace{-.25cm}
\end{table}

\textbf{Tidal height, wind speed, and air temperature data set}.\footnote{
    The data can be downloaded at \url{http://www.bramblemet.co.uk}, \url{http://cambermet.co.uk}, \url{http://www.chimet.co.uk}, and \url{http://sotonmet.co.uk}.
}
This data set was collected at 5 minute intervals by four weather stations: Bramblemet, Cambermet, Chimet, and Sotonmet, all located in Southampton, UK.
The task is to predict the air temperature measured by Cambermet and Chimet from all other signals.
Performance is measured with the SMSE.
This experiment serves two purposes. First, it demonstrates that it is simple to scale GPAR to large data sets using off-the-shelf inducing point techniques for single-output GP regression. Second, it shows that scaling to large data sets enables GPAR to better learn dependencies between outputs, which, importantly, can significantly improve predictions in regions where outputs are partially observed.
We utilise the variational inducing point method by \citet{Titsias:2009:Variational_Learning} as discussed in \cref{sec:gpar}, with 10 inducing points per day.
This data set is not closed downwards, so we use mean imputation when training.
We use D-GPAR-L and set the temporal kernel to be a simple EQ, meaning that the model \textit{cannot} make long-range predictions, but instead must exploit correlations between outputs.

\citet{Nguyen:2014:Collaborative_Multi-Output} consider from this data set 5 days in July 2013, and predict short periods of the air temperature measured by Cambermet and Chimet using all other signals.
We followed their setup and predicted the same test set, but instead trained on the \textit{whole} of July.
Even though the additional observations do not temporally correlate with the test periods at all, they enable GPAR to better learn the relationships between the outputs, which, unsurprisingly, significantly improved the predictions: using the whole of July, GPAR achieves SMSE $0.056$, compared to their SMSE $0.107$.

\begin{figure*}[t]
    \centering
    \centerline{\includegraphics[width=\linewidth]{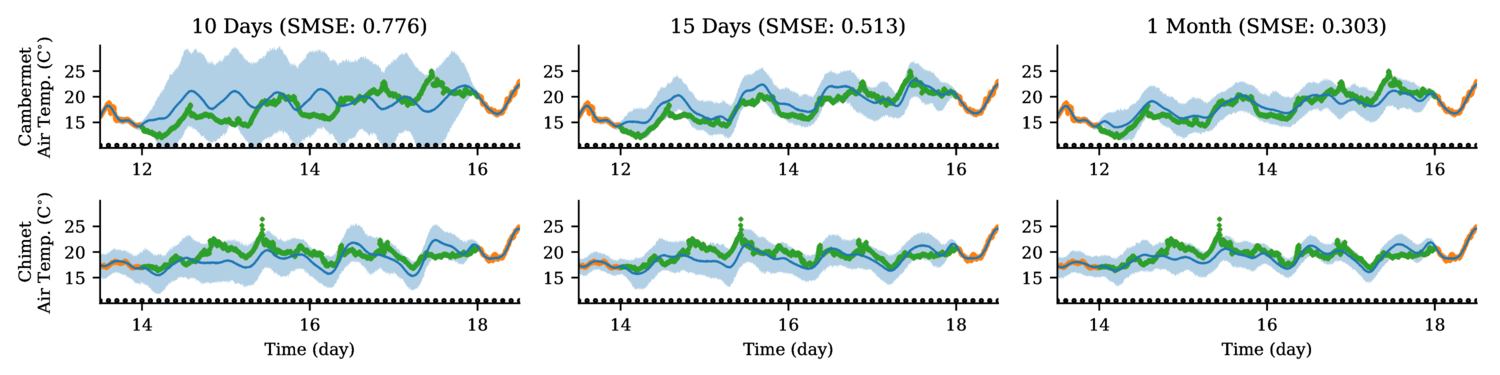}}
    \vspace{-1em}
    \caption{Visualisation of the air temperature data set and GPAR's prediction for it. Black circles indicate the locations of the inducing points.}
    \label{fig:air_prediction}
    \vspace{-.5cm}
\end{figure*}

The test set used by \citet{Nguyen:2014:Collaborative_Multi-Output} is rather small, yielding high-variance test results.
We therefore do not pursue further comparisons on their train--test split, but instead consider a bigger, more challenging setup:
using as training data 10 days (days $[10,20)$, roughly \SI{30}{k} points), 15 days (days $[18,23)$, roughly \SI{47}{k} points), and the whole of July (roughly \SI{98}{k} points), make predictions of 4 day periods of the air temperature measured by Cambermet and Chimet.
\Cref{fig:air_prediction} visualises the test periods and GPAR's predictions for it.
Despite the additional observations not correlating with the test periods, we observe clear, though dimishining, improvements in the predictions as the training data is increased.

\begin{figure}[t]
    \centering
    \centerline{\includegraphics[width=\linewidth]{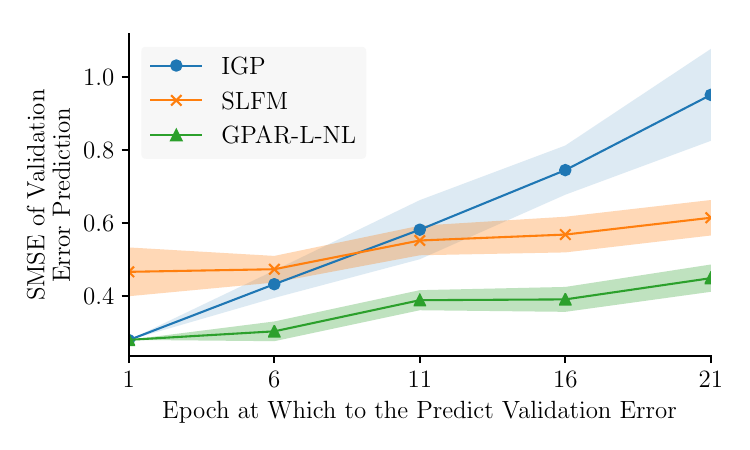}}
    \vspace{-1em}
    \caption{Results for the machine learning data set for a GP, the SLFM with two latent dimensions, and GPAR}
    \label{fig:ml_results}
    \vspace{-.75cm}
\end{figure}

\textbf{MLP validation error data set}. The final data set is the validation error of a multi-layer perceptron (MLP) on the MNIST data, trained using categorical cross-entropy, and set as a function of six hyperparameters: the number of hidden layers, the number of neurons per hidden layer, the dropout rate, the learning rate to use with the ADAM optimizer, the $L_1$ weight penalty, and the $L_2$ weight penalty. This experiment was implemented using code made available by \citet{jmhl2016blog}. An improved model for the objective surface could translate directly into improved performance in Bayesian optimisation \citep{snoek2012practical}, as this would enable a more informed search of the hyperparameter space.

To generate a data set, we sample 291 sets of hyperparameters randomly from a rectilinear grid and train the MLP for 21 epochs under each set of hyperparameters, recording the validation performance after 1, 5, 11, 16, and 21 epochs. We construct a training set of 175 of these hyperparameter settings and, crucially, discard roughly 30\% of the validation performance results at 5 epochs at random, and again discard roughly 30\% of those results at 11 epochs, and so forth. The resulting data set has 175 labels after 1 epoch, 124 after 5, 88 after 11, 64 after 15 and 44 after 21, simulating the partial completion of the majority of runs. Importantly, a Bayesian Optimisation system typically exploits only completed training runs to inform the objective surface, whereas GPAR can also exploit partially complete runs.

The results presented in \cref{fig:ml_results} show the SMSE in predicting validation performance at each epoch using GPAR, the SLFM, and independent GPs on the test set, averaged over 10 seed for the pseudo-random number generator used to select which outputs from the training set to discard.
GPs trained independently to predict performance after a particular number of epochs perform worse than the SLFM and GPAR, which both have learned to exploit the extra information available.
In turn, GPAR noticeably outperforms the SLFM.

\section{Conclusion and Future Work}
\label{sec:conclusion}
This paper introduced GPAR: a flexible, fast, tractable, and interpretable approach to multi-output GP regression. GPAR can model (1) nonlinear relationships between outputs and (2) complex output noise.
GPAR can scale to large data sets by trivially leveraging scaling techniques for one-dimensional GP regression \citep{Titsias:2009:Variational_Learning}.
In effect, GPAR transforms high-dimensional data modelling problems into set of single-output modelling problems, which are the bread and butter of the GP approach.
GPAR was rigorously tested on a variety of synthetic and real-world problems, consistently outperforming existing GP models for multi-output regression.
An exciting future application of GPAR is to use compositional kernel search \citep{LloDuvGroetal2014} to automatically learn and explain dependencies between outputs and inputs.
Further insights into structure of the data could be gained by decomposing GPAR's posterior over additive kernel components \citep{Duvenaud:2014:Automatic_Construction}.
These two approaches could be developed into a useful tool for automatic structure discovery.
Two further exciting future applications of GPAR are modelling of environmental phenomena and improving data efficiency of existing Bayesian optimisation tools \citep{snoek2012practical}.

\clearpage

\section*{Acknowledgements}
Richard E. Turner is supported by Google as well as EPSRC grants EP/M0269571 and EP/L000776/1.

\bibliographystyle{apalike}
\bibliography{bibliography.bib}

\clearpage
\appendix

\section{Conditional Independence in Figure 2b}
\label{app:conditional_independence}

In this section, we prove the key conditional independence in Figure 2b that makes GPAR work:

\begin{theorem} \label{thm:d-separation}
    Let a set of observations $\mathcal{D}$ be closed downwards.
    Then $y_{i} \perp \mathcal{D}_{i+1:M} \cond \mathcal{D}_{1:i}$, where $\mathcal{D}_{1:i}$ are the observations for outputs $1,\ldots,i$ and $\mathcal{D}_{i+1:M}$ for $i+1,\ldots,M$.\footnote{
        $\mathcal{D}_{1:i}=\set{y_j^{(n)} \in \mathcal{D}: j \le i,\, n \le N}$ and 
        $\mathcal{D}_{i+1:M}=\set{y_j^{(n)} \in \mathcal{D}: j > i,\, n \le N}$.
        }
\end{theorem}

To begin with, we review some basic notions concerning graphical models.
Let a path be a sequence of nodes $v_1, \ldots, v_n$ from some directed graph $G$ where, for each
$v_i$, $G$ either contains an edge from $v_i$ to $v_{i+1}$, or an edge from $v_{i+1}$ to $v_i$.
If $G$ contains an edge from $a$ to $b$,
then write $a \to b$ to mean the two-node path in which
node $b$ follows node $a$.
Similarly, if $G$ contains an edge from $b$ to $a$,
then write $a \gets b$ to mean the two-node path in which
$b$ follows $a$. Write $a \rlh b$ to mean either $a \to b$ or $b \gets a$.

\begin{definition}[Active Path (Definition 3.6 from \citet{Koller:2009:Probabilistic_Graphical_Models_Principles_and})]
    Let $\mathcal{P} = v_1 \rlh \cdots \rlh v_n$ be a path in a graphical model. Let $Z$ be
    a subset of the variables from the graphical model. Then, call $\mathcal{P}$ active given $Z$ if
    (1) for every v-structure $v_{i-1} \to v_i \gets v_{i+1}$ in $\mathcal{P}$, $v_i$ or a descendant of $v_i$ is in $Z$; and
    (2) no other node in $\mathcal{P}$ is in $Z$.
\end{definition}

\begin{definition}[d-Separation (Definition 3.7 from \citet{Koller:2009:Probabilistic_Graphical_Models_Principles_and})]
    Let $X$, $Y$, and $Z$ be three sets of nodes from a graphical model. Then, call $X$ and $Y$ d-separated given $Z$ if no path between any $x \in X$ and $y \in Y$ is active given $Z$.
\end{definition}

\begin{theorem}[d-Separation Implies Conditional Independence (Theorem 3.3 from \citet{Koller:2009:Probabilistic_Graphical_Models_Principles_and})]
    Let $X$, $Y$, and $Z$ be three sets of nodes from a graphical model. If $X$ and $Y$ are d-separated given $Z$, then $X \perp Y \cond Z$.
\end{theorem}

Define the \textit{layer} of a node in Figure 2b to be
\newcommand{\layer}{\operatorname{layer}}
\begin{align*}
    \layer(f_i) = \layer(y_i(\vec{x})) = i.
\end{align*}
We are now ready to prove \cref{thm:d-separation}.

\begin{proof}[Proof of \cref{thm:d-separation}]
    For $i < j$, let $\mathcal{P}$ be a path between any $y_i(\vec{x}') \in y_i$ and $y_j(\vec{x}) \in \mathcal{D}_{i+1:N}$.
    Let $y_k(\hat{x})$ be the first node in $\mathcal{P}$ such that $\layer(y_k(\hat{x})) > i$.
    Then $\mathcal{P}$ contains
    \begin{align*}
        \cdots \to y_m(\hat{\vec{x}}) \to y_k(\hat{\vec{x}}) \rlh \cdots
    \end{align*}
    for some $m \le i < k$.
    
    If $y_k(\hat{\vec{x}}) \in \mathcal{D}_{i+1:N}$, then, since $\mathcal{D}$ is closed downwards, $y_m(\hat{\vec{x}}) \in \mathcal{D}_{1:i}$, meaning that $\mathcal{P}$ is inactive.
    
    If, on the other hand, $y_k(\hat{\vec{x}}) \notin \mathcal{D}$, then, since $\mathcal{D}$ is closed downwards, $y_{k'}(\hat{x}) \notin \mathcal{D}$ for all $k' \ge k$.
    Therefore, $y_j(x)$ cannot be descendant of $y_k(\hat{x})$, so $\mathcal{P}$ must contain
    \begin{align*}
        \cdots \to y_{k'}(\hat{x}) \to y_{k''}(\hat{x}) \gets f_{k''} \to \cdots
    \end{align*}
    for some $m \le k' < k''$, which forms a v-structure.
    We conclude that $\mathcal{P}$ is inactive, because $y_{k''}(\hat{x})$ nor a descendant of $y_{k''}(\hat{x})$ can be in $\mathcal{D}$.
\end{proof}

\section{The Nonlinear and Linear Equivalent Model}
\label{app:equivalent_models}

In this section, we construct equivalent models for GPAR (\cref{lem:nonlinear_equivalent_model,lem:linear_equivalent_model}).
These models make GPAR's connection to other models in the literature explicit.

To begin with, we must introduce some notation and definitions.
For functions
% $\vec{A},\vec{B}\colon \mathcal{X} \times (\mathcal{Y}^\mathcal{X})^M \to \mathcal{Y}^M$,
$\vec{A},\vec{B}\colon \mathcal{X} \times (\mathcal{Y}^M)^\mathcal{X} \to \mathcal{Y}^M$,
define composition $\circ$ as follows: $(\vec{A} \circ \vec{B})(\vec{x},\vec{y}) = \vec{A}(\vec{x},\vec{B}(\vardot,\vec{y}))$. Note that $\circ$ is well-defined and associative. For a function $\vec{u}\colon \mathcal{X} \to \mathcal{Y}^M$, denote $\vec{A} \circ \vec{u} \colon \mathcal{X} \to \mathcal{Y}^M$, $\vec{A}\circ \vec{u} = \vec{A}(\vardot,\vec{u})$. Again, note that $(\vec{A} \circ \vec{B})\circ \vec{u} = \vec{A} \circ (\vec{B} \circ \vec{u})$. Furthermore, denote
\begin{align*}
    \underbrace{\vec{A} \circ \cdots \circ \vec{A}}_{n\text{ times}} = \vec{A}^n.
\end{align*}

Consider a function $\vec{A}\colon \mathcal{X} \times (\mathcal{Y}^M)^\mathcal{X} \to \mathcal{Y}^M$ such that $A_i(\vec{x},\vec{y})\colon \mathcal{X} \times (\mathcal{Y}^M)^\mathcal{X} \to \mathcal{Y}$ depends only on $(\vec{x},\vec{y}_{1:i-1})$, where $A_1=0$. Further let $\vec{u},\vec{y}\colon \mathcal{X} \to \mathcal{Y}^M$, denote $\Top \vec{f} = \vec{u} + \vec{A} \circ \vec{f}$, and denote $N$ consecutive applications of $\Top$ by $\Top^N$.

The expression $\Top^{M-1} \vec{u}$ will be key in constructing the equivalent models.
We show that it is the unique solution of a functional equation:

\begin{proposition} \label{lem:fp_solution}
     The unique solution of $\vec{y} = \vec{u} + \vec{A} \circ \vec{y}$ is $\vec{y} = \Top^{M-1} \vec{u}$.
\end{proposition}

\begin{proof}[Proof of \cref{lem:fp_solution}]
    First, we show that $\vec{y} = \vec{u} + \vec{A} \circ \vec{y}$ has a solution, and that this solution is unique. Because $A_i(\vec{x},\vec{y})$ depends only on $(\vec{x},\vec{y}_{1:i-1})$, it holds that
    \begin{align*}
        y_i = u_i + A_i \circ \vec{y} = u_i + A_i \circ (\vec{y}_{1:i-1},\vec{0}),
    \end{align*}
    where $(\vec{y}_{1:i-1},\vec{0})$ represents the concatenation of $\vec{y}_{1:i-1}$ and $M-i+1$ zeros.
    Thus, $y_i$ can uniquely be constructed from $u_i$, $A_i$, and $\vec{y}_{1:i-1}$; therefore, $y_1$ exists and is unique, so $y_2$ exists and is unique: by induction we find that $\vec{y}$ exists and is unique.

    Second, we show that $\vec{y} = \Top^{M-1} \vec{u}$ satisfies $\vec{y} = \vec{u} + \mat{A} \circ \vec{y}=\Top \vec{y}$. To show this, we show that $(\Top^n \vec{u})_{i} = (\Top^{n-1} \vec{u})_{i}$ for $i=1,\ldots,n$, for all $n$. To begin with, we show the base case, $n=1$:
    \begin{align*}
        (\Top\vec{u})_1 = u_1 + A_1 \circ \vec{u} = u_1 = (\Top^0\vec{u})_1,
    \end{align*}
    since $A_1=0$. Finally, suppose that the claim holds for a particular $n$. We show that the claim then holds for $n+1$: Let $i \le n+1$. Then
    \begin{align*}
        (\Top^{n+1} \vec{u})_{i}
        &= u_i + A_i\circ \Top^n \vec{u} \\
        &= u_i + A_i \circ ((\Top^n \vec{u})_{1:i-1},(\Top^n \vec{u})_{i:M}) \\
        &= u_i + A_i \circ ((\Top^{n-1} \vec{u})_{1:i-1},(\Top^n \vec{u})_{i:M})
        \tag{\text{By assumption}} \\
        &\overset{\smash{\mathclap{(\ast)}}}{=} u_i + A_i \circ ((\Top^{n-1} \vec{u})_{1:i-1},(\Top^{n-1} \vec{u})_{i:M}) \\
        &= u_i + A_i\circ \Top^{n-1} \vec{u} \\
        &=(\Top^{n} \vec{u})_{i},
    \end{align*}
    where $(\ast)$ holds because $A_i(\vec{x},\vec{y})$ depends only on $(\vec{x},\vec{y}_{1:i-1})$.
\end{proof}

In the linear case, $\Top^{M-1}\vec{u}$ turns out to greatly simplify.

\begin{proposition}\label{lem:fp_solution_linear}
    If $\vec{A}(\vec{x},\vec{y})$ is linear in $\vec{y}$, then $\Top^{M-1}\vec{u} = (\sum_{i=0}^{M-1}\vec{A}^i)\circ\vec{u}$.
\end{proposition}
\begin{proof}[Proof of \cref{lem:fp_solution_linear}]
    If $\vec{A}(\vec{x},\vec{y})$ is linear in $\vec{y}$, then one verifies that $\circ$ distributes over addition.
    Therefore,
    \begin{align*}
        \Top^{M-1}\vec{u}
        &= \vec{u} + \vec{A} \circ \Top^{M-2}\vec{u} \\
        &= \vec{u} + \vec{A} \circ \vec{u} + \vec{A}^2 \circ \Top^{M-3}\vec{u} \\
        &\;\;\vdots \\
        &= \vec{u} + \vec{A} \circ \vec{u} + \cdots + \vec{A}^{M-1} \circ \vec{u}. \qedhere
    \end{align*}
\end{proof}

We now use \cref{lem:fp_solution,lem:fp_solution_linear} to construct a nonlinear and linear equivalent model.

\begin{lemma}[Nonlinear Equivalent Model] \label{lem:nonlinear_equivalent_model}
    Let $\mat{A}$ be an $M$-dimensional vector-valued process over $\mathcal{X}\times (\mathcal{Y}^M)^\mathcal{X}$, each $A_{i}$ drawn from $\mathcal{GP}(0, k_{A_i})$ independently, and let $\vec{u}$ be an $M$-dimensional vector-valued process over $\mathcal{X}$, each $u_i$ drawn from $\mathcal{GP}(0, k_{u_i})$ independently. Furthermore, let $A_{i}(\vec{x},\vec{y})\colon \mathcal{X}\times (\mathcal{Y}^M)^\mathcal{X}\to \mathcal{Y}$ depend only on $(\vec{x},\vec{y}_{1:i-1})$, meaning that $k_{A_i}=k_{A_i}(\vec{x},\vec{y}_{1:i-1},\vec{x}',\vec{y}_{1:i-1}')$, and let $A_1=0$. Denote $\Top \vec{f} = \vec{u} + \vec{A} \circ \vec{f}$, and denote $N$ consecutive applications of $\Top$ by $\Top^N$. Then
    \begin{gather*}
        \vec{y}\cond \mat{A}, \vec{u} = \Top^{M-1} \vec{u}
        \iff \\
        y_{i}\cond \vec{y}_{1:i-1} \sim \mathcal{GP}(0, k_{u_i} + k_{A_i}(\vardot, \vec{y}_{1:i-1},\vardot,\vec{y}_{1:i-1})).
    \end{gather*}
\end{lemma}
\begin{proof}[Proof of \cref{lem:nonlinear_equivalent_model}]
    Since $A_i(\vec{x},\vec{y})$ depends only on $(\vec{x},\vec{y}_{1:i-1})$, it holds by \cref{lem:fp_solution} that any sample from $\vec{y}\cond \mat{A}, \vec{u}$ satisfies $y_i = u_i + A_i \circ \vec{y}$, so $y_i = u_i + A_i \circ (\vec{y}_{1:i-1},\vec{0})$, where $(\vec{y}_{1:i-1},\vec{0})$ represents the concatenation of $\vec{y}_{1:i-1}$ and $M-i+1$ zeros. The equivalence now follows.
\end{proof}

\begin{lemma}[Linear Equivalent Model] \label{lem:linear_equivalent_model}
    Suppose that $\vec{A}$ was instead generated from
    \begin{align*}
        \vec{A}(\vec{x},\vec{y})\cond\hat{\mat{A}}=\int\hat{\mat{A}}(\vec{x}-\vec{z})\vec{y}(\vec{z})\sd{\vec{z}},
    \end{align*}
    where $\hat{\mat{A}}$ is an $(M\times M)$-matrix-valued process over $\mathcal{X}$, each $\hat{A}_{i,j}$ drawn from $\mathcal{GP}(0,k_{\hat{A}_{i,j}})$ independently if $i > j$ and $\smash{\hat{A}}_{i,j}=0$ otherwise. Then
    \begin{gather}
        \vec{y}\cond \mat{A}, \vec{u} = \left(\sum_{i=0}^{M-1}\vec{A}^i\right)\circ\vec{u} \iff \nonumber \\
        y_{i}\cond \vec{y}_{1:i-1} \sim \mathcal{GP}(0, k_{u_i} + k_{A_i}(\vardot, \vec{y}_{1:i-1},\vardot,\vec{y}_{1:i-1})), \label{eq:linear_equivalent_model_equivalence}
    \end{gather}
    where
    \begin{align*}
        &k_{A_i}(\vec{x}, \vec{y}_{1:i-1},\vec{x}',\vec{y}_{1:i-1}') \\
        &\quad= \sum_{j=1}^{i-1} \int k_{\hat{A}_{i,j}}(\vec{x}-\vec{z},\vec{x}'-\vec{z}')y_j(\vec{z})y'_j(\vec{z}')\sd{\vec{z}}\sd{\vec{z}'}.
    \end{align*}
\end{lemma}
\begin{proof}[Proof of \cref{lem:linear_equivalent_model}]
    First, one verifies that $A_i(\vec{x},\vec{y})$ still depends only on $(\vec{x},\vec{y}_{1:i-1})$, and that $A_i(\vec{x},\vec{y})$ is linear in $\vec{y}$. The result then follows from \cref{lem:nonlinear_equivalent_model,lem:fp_solution_linear}, where the expression for $k_{A_i}$ follows from straightfoward calculation.
\end{proof}

As mentioned in the paper, the kernels for $f_{1:M}$ determine the types of relationships between inputs and outputs that can be learned. \Cref{lem:nonlinear_equivalent_model,lem:linear_equivalent_model} make this explicit: \cref{lem:nonlinear_equivalent_model} shows that nonlocal GPAR can recover a model where $M$ latent GPs $\vec{u}$ are repeatedly composed with another latent GP $\mat{A}$, where $\mat{A}$ has a particular dependency structure, and \cref{lem:linear_equivalent_model} shows that nonlocal GPAR can recover a model where $M$ latent GPs $\vec{u}$ are linearly transformed, where the linear transform $\mat{T}=\sum_{i=0}^{M-1}\mat{A}^i$ is lower triangular and may vary with the input.

In \cref{lem:linear_equivalent_model}, note that it is not restrictive that $\mat{T}$ is lower triangular: Suppose that $\mat{T}$ were dense. Then, letting $\vec{y}\cond \mat{T}, \vec{u}=\mat{T}\circ\vec{u}$, $\vec{y}\cond\mat{T}$ is jointly Gaussian. Hence $y_{i}\cond \vec{y}_{1:i-1},\mat{T}$ is a GP whose mean linearly depends upon $\vec{y}_{1:i-1}$ via $\mat{T}$, meaning that $y_{i}\cond \vec{y}_{1:i-1}$ is of the form of \cref{eq:linear_equivalent_model_equivalence} where $k_{u_i}$ may be more complicated. If, however, $\hat{\vec{A}}(\vec{z}) = \delta(\vec{z})\vec{B}$ for some random $(M \times M)$-matrix $\vec{B}$, each $B_{i,j}$ drawn from $\mathcal{N}(0, \sigma^2_{B_{i,j}})$ if $i>j$ and $B_{i,j}=0$ otherwise, then it is restrictive that $\mat{T}$ is lower triangular: In this case, $\vec{y}(\vec{x})\cond \mat{B},\vec{u} = \sum_{i=0}^{M-1} \mat{B}^i \vec{u}(\vec{x})$. If $\mat{T}=\sum_{i=0}^{M-1} \mat{B}^i$ were dense, then, letting $\vec{y}\cond \mat{T}, \vec{u}=\mat{T}\vec{u}$, $\vec{y}$ can be represented with \cref{lem:linear_equivalent_model} if and only if $\vec{y}\cond \mat{T}$'s covariance can be diagonalised by a constant, invertible, lower-triangular matrix. This condition does not hold in general, as \cref{lem:triangular} proves.

\section{Lemma \ref{lem:triangular}}
\label{app:triangular}
Call functions $k_1,\ldots,k_M\colon \mathcal{X} \to \R$ linearly independent if
\begin{align*}
    \left( \forall\, \vec{x} : \sum_{i=1}^M c_i k_i(\vec{x})=0 \right) \implies c_1=\ldots=c_M=0.
\end{align*}

\begin{lemma} \label{lem:triangular}
    Let $k_1,\ldots,k_M\colon \mathcal{X} \to \R$ be linearly independent and arrange them in a diagonal matrix ${\mat{K}=\diag (k_1,\ldots, k_n)}$. Let $\mat{A}$ be an invertible $M\times M$ matrix such that its columns cannot be permuted into a triangular matrix. Then there does not exist an invertible triangular matrix $\mat{T}$ such that $\mat{T}^{-1}\mat{B} \mat{K}(\vec{x}) \mat{B}^\T \mat{T}^{-\T}$ is diagonal for all $\vec{x}$.
\end{lemma}
\begin{proof}
    Suppose, on the contrary, that such $\mat{T}$ does exist. Then two different rows $\vec{a}_p$ and $\vec{a}_q$ of $\mat{A}=\mat{T}^{-1}\mat{B}$ share nonzero elements in some columns $C$; otherwise, $\mat{A}$ would have exactly one nonzero entry in every column---$\mat{A}$ is invertible---so $\mat{A}$ would be the product of a permutation matrix and a diagonal matrix, meaning that $\mat{B}=\mat{T}\mat{A}$'s columns could be permuted into a triangular matrix. Now, by $\mat{T}^{-1}\mat{B} \mat{K}(\vec{x}) \mat{B}^\T \mat{T}^{-\T}=\mat{A} \mat{K}(\vec{x}) \mat{A}^\T$ being diagonal for all $\vec{x} \in \mathcal{X}$, $\sum_{i} \vec{a}_{p,i} \vec{a}_{q,i} k_i(\vec{x})=0$ for all $\vec{x}$. Therefore, by linear independence of $k_1,\ldots, k_N$, it holds that $\vec{a}_{p,i}\vec{a}_{q,i}=0$ for all $i$. But $\vec{a}_{p,i}\vec{a}_{q,i} \neq 0$ for any $i \in C$, which is a contradiction.
\end{proof}

\section{Experimental Details}
For every experiment, the form of the kernels is determined by the particular GPAR model used: GPAR-L, GPAR-NL, or GPAR-L-NL (see Table 2 in the main paper), potentially with a D-$\ast$ prefix to indicate the denoising procedure outlined in ``Potential deficiencies of GPAR'' in Section 2 of the paper main.
For GPAR-NL, we always used exponentiated quadratic (EQ) kernels, except for the exchange rates experiment, where we used rational quadratic (RQ) kernels \citep{Rasmussen:2006:Gaussian_Processes}.
Furthermore, in every problem we simply expanded according to Equation (1) in the main paper or greedily optimised the ordering, in both cases putting the to-be-predicted outputs last.
We used \texttt{scipy}'s implementation of the L-BFGS-B algorithm \citep{Nocedal:2006:Numerical_Optimisation} to optimise hyperparameters.

\end{document}